%% file: main.tex
\documentclass{article}

\usepackage{neurips_2018_like}

\usepackage[utf8]{inputenc} 
\usepackage[T1]{fontenc}    
\usepackage{hyperref}       
\usepackage{url}            
\usepackage{booktabs}       
\usepackage{tabularx}
\usepackage{ltxtable}
\usepackage{amsfonts}       
\usepackage{nicefrac}       
\usepackage{microtype}      
\usepackage{graphicx}
\usepackage{mathtools}
\usepackage{gensymb}
\usepackage{caption} 
\usepackage[title]{appendix}
\usepackage{float}

\title{Deep Recurrent Q-Learning vs Deep Q-Learning on a simple Partially Observable Markov Decision Process with Minecraft}

\author{%
  Clément Romac \\
  Ynov Informatique\\
  Bordeaux, FR \\
  \texttt{clement.romac@ynov.com} \\
	\And
	Vincent Béraud \\
  Ynov Informatique\\
  Bordeaux, FR \\
  \texttt{vincent.beraud@ynov.com} \\
}

\begin{document}

\maketitle

\begin{abstract}
	Deep Q-Learning has been successfully applied to a wide variety of tasks in the past several years. However, the architecture of the vanilla Deep Q-Network is not suited to deal with partially observable environments such as 3D video games. For this, recurrent layers have been added to the Deep Q-Network in order to allow it to handle past dependencies. We here use Minecraft for its customization advantages and design two very simple missions that can be frames as Partially Observable Markov Decision Process. We compare on these missions the Deep Q-Network and the Deep Recurrent Q-Network in order to see if the latter, which is trickier and longer to train, is always the best architecture when the agent has to deal with partial observability.
\end{abstract}

\section{Introduction}
	Deep Reinforcement Learning has been highly active since the successful work of \citet{mnih_playing_atari_2013} on Atari 2600 games. From that moment, a lot of methods have been used on a wide range of environments in order to make an agent reach an objective \citep{justesen_deep_2017}. These environments can be framed as Markov Decision Processes (MDPs) defined by the tuple \(<S, A, P, R>\) where at each timestep \(t\) of the environment, the agent observes a state \(s \in{S}\), takes the action \(a \in{A}\), ends up in a new state \(s' \sim P(s, a)\) and receives a reward \(r \in{R}\).
	
	In order to find the policy \(\pi\) (the choice of the action from the state) maximizing the sum of rewards, one can use the Q-Learning algorithm \citep{watkins_q-learning_1992} to estimate the expected sum of rewards from a state \(s\) and an action \(a\). This expected sum of reward is called the Q-value and is noted as : \[Q(s, a) = \mathbb{E}[R_{t+1} + \gamma Q(S_{t+1}, A_{t+1})| S_t = s, A_t = A].\] 
	This is actually a discounted sum with \(\gamma \in [0, 1]\). This discounting factor allows us to handle how important future rewards are, but also prevents the algorithm from an infinite loop. Although useful, finding the Q-value with the Q-Learning algorithm for all the state-action pairs is simply intractable on complex environments involving wide states and actions spaces.
	
	In order to solve this issue, we can approximate this Q-value with a neural network  (called Q-Network) which takes the state in input and outputs the approximated Q-value for each possible action. This technique is called Deep Q-Learning \citep{mnih_playing_atari_2013} and has moved the state of the art on several Atari games to a human-level performance \citep{mnih_human-level_2015}. In the context of vision-based environments, these Deep Q-Networks usually take in input the raw image of the current state and use a convolutional architecture to extract feature maps. These feature maps then go to fully-connected layers with the last one outputting the approximated Q-values. The fact of taking only the current observation as an image in input and choosing the action knowing only that information makes the assumption that the observation holds all the knowledge about the current state of the environment. This is unfortunately not always the case. Let's take for instance the Atari game Pong. The image gives us the information about the location of the ball but doesn't give us neither the information about the trajectory nor the speed. This is also the case for 3D environments such as Doom (a first-person shooter game).
	
	These environments where an observation gives us only a partial information about the current state of the game can be framed as Partially Observable Markov Decision Processes (POMDPs) defined by the tuple \(<S, A, P, R, \Omega, O>\) where \(S\), \(A\), \(P\) and \(R\) are the same as the MDP. The main difference is that the agent no longer observes the whole current state but instead receives an observation \(o \in \Omega\) from the probability distribution \(o \sim O(s, a)\). In order to find the best policy \(\pi\) for this POMDP, we can still use the Q-Learning algorithm and approximate the Q-value with a neural network. However, the Deep Q-Networks aforementioned are not meant to deal with this partial observability as the network needs more than one observation to understand the current state.
	
	In practice, \citet{mnih_playing_atari_2013} used the last four frames as input of their DQN to deal with partially observable environments such as Pong. But the use of stacked frames has drawbacks. The number of stacked frames is fixed and our network can't handle older things than this fixed size. Some complex environments may need the agent to remember events that occurred tens of frames before. In order to fix this memory issue of the DQN, \citet{hausknecht_deep_2015} proposed to add a recurrent layer after the convolutional part of the original DQN. They used LSTM cells \citep{hochreiter_long_1997} to handle long term dependencies and called this architecture the Deep Recurrent Q-Network (DRQN). They showed that the DRQN could outperform the vanilla DQN on some of the Atari 2600 games. \citet{lample_playing_2016} also showed that the DRQN architecture could achieve good results on complex 3D environments such as Doom.
	
	\begin{figure}[hbt!]
        \centering
        \includegraphics[width=200pt]{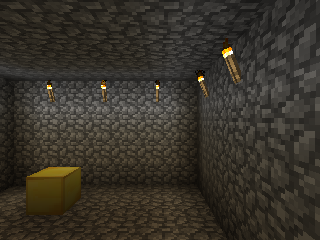}
        \caption{An overview of Minecraft in the Basic environment (taken from gym-minecraft's documentation).}
        \label{fig:minecraft_overview}
    \end{figure}
	
	In this paper, we want to check if using a Deep Recurrent Q-Network, which is trickier and longer to train than the classical DQN, is always the best choice when we are in a POMDP, even if the mission to solve is simple. We thus chose a 3D environment called Minecraft, which is a sandbox video game where an agent can live, build, and explore an open world. With the help of the Project Malmö\footnote{https://www.microsoft.com/en-us/research/project/project-malmo/} \citep{johnson_malmo_2016} and Gym-Minecraft\footnote{https://github.com/tambetm/gym-minecraft}, we were able to design missions for an agent that can be solved with Deep Reinforcement Learning. This environment has the advantage of being highly customizable.
	
	In order to see if a modification of the network architecture is always needed to deal with the partial observability, we compare the performances of three models :
	\begin{itemize}
	    \item A Deep Q-Network constituted of a convolutional neural network that takes in input only the current frame. 
        \item The \citet{mnih_playing_atari_2013} Deep Q-network which is similar to the previous one but takes the last four frames in input.
        \item A Deep Recurrent Q-Network with a convolutional part taking the current frame as input, then followed by an LSTM before the feedforward layers. 
    \end{itemize}
    
    We designed two simple environments resolvable by a vanilla DQN :
	\begin{itemize}
        \item The Basic : the agent is in a 7x7x7 blocks closed room where a goal has been placed randomly on one of the axis (see figure \ref{fig:minecraft_overview}).
        \item The Cliff Walking : the agent has to walk through a pathway of blocks surrounded by lava to reach is goal.
    \end{itemize}
    We measure the performances of the models on both the training and evaluation phases. Our implementations can be found on our GitHub repository\footnote{https://github.com/vincentberaud/Minecraft-Reinforcement-Learning}. 
    
\section{Related Work} \label{related_work}
    Some works have already tried to apply Reinforcement Learning on Minecraft. \citet{alaniz_deep_2018} used Minecraft to make their agent place a cube at the right place. They chose to use a model-based approach where a convolutional neural network (called the Transition Model) takes as input the last four frames and outputs the next frame and the associated reward. They then used the Monte Carlo Tree Search algorithm to do planning and choose the action that would maximize the expected sum of rewards based on the Transition Model predictions. Their results showed that their method seems to learn faster than the DQN but also seems to be less efficient as the number of training steps increases.
    
    \citet{oh_control_2016} had an approach closer to ours. They used Minecraft to make an agent resolve a maze where there are two ending paths with an indicator (at the beginning of the maze) showing the right one to choose. Their agent uses a Deep Recurrent Q-Network but with an external memory and a feedback connection. They called their architecture the Feedback Recurrent Memory Q-Network (FRMQN). Their method outperformed both the DRQN and the DQN in their environment.
    
    Finally, \citet{matiisen_teacher-student_2017} made their agent find a goal in a room with hierarchical tasks by using a transfer learning technique between a neural network (Teacher) with a partial view (POMDP) and an agent which is another neural network (Student) performing tasks more and more complex. They called their approach Teacher-Student Curriculum Learning (TSCL) and used the Proximal Policy Optimization (PPO) algorithm \citep{schulman_proximal_2017} to learn the policy. Our use of Gym-Minecraft is inspired by their work and our basic mission is the first mission their agent had to accomplish.
    
    We choose here to highlight the behaviour of the DRQN versus two DQNs versions (one handling some partial observability and not the other one) on simpler environments than the ones just evoked in order to have a baseline of the need of the DRQN when it comes to partial observability. 

\section{Methods}
    \subsection{Q-Learning}
        In Reinforcement Learning, an agent has to learn the policy \(\pi\) that maximizes the sum of discounted future rewards. This sum is called the \textbf{return} and is noted as :
        \[G_t = R_{t+1} + \gamma R_{t+2} + \gamma^2 R_{t+3} + ... = \sum_{k=0}^{\infty}\gamma^k R_{t+k+1}.\]
        Knowing this, we can say that the "quality" of a state (how rewarding it is to be in this state) is the return of following the policy $\pi$ starting from that state $s$ . This is called the  \textbf{value} and is written as :
        \[V(s)_{\pi} = \mathbb{E}_{\pi}[G_t|S_t = s].\]
        We can now be even more precise and write how good it is of being in the state $s$ and choose the action $a$ from that state. This is called the \textbf{Q-value} and can thus be written as :
        \begin{equation}
            \begin{split}
                Q(s,a)_{\pi} & = \mathbb{E}_{\pi}[G_t|S_t = s, A_t = a] \\
                             & = \mathbb{E}_{\pi}[R_{t+1} + \gamma G_{t+1} | S_t = s, A_t = a] \\
                             & = \mathbb{E}_{\pi}[R_{t+1} + \gamma Q(S_{t+1}, A_{t+1} | S_t = s, A_t = a]
            \end{split}
        \end{equation}
        In order to find the optimal policy (noted $\pi_*$), we want to find the optimal Q-value :
        \begin{equation}
            \begin{split}
                Q_{\pi_*}(s,a) & = \max_{\pi}Q_{\pi}(s,a) \\
                              & = Q_*(s,a)
            \end{split}
        \end{equation}
        
        In order to do so, we can use the Q-Learning \citep{watkins_q-learning_1992} algorithm. Q-Learning is a model-free off-policy method that estimates the Q-value using Temporal Difference (TD) Learning \citep{sutton_learning_1988}. The key idea behind it, is that it updates iteratively the estimated Q-value towards the true Q-value :
        \begin{equation}
            \begin{split}
                Q(S_t,A_t) & \leftarrow (1-\alpha)Q(S_t,A_t) + \alpha G_t \\
                           & \leftarrow Q(S_t, A_t) + \alpha(G_t - Q(S_t, A_t)) \\
                           & \leftarrow Q(S_t, A_t) + \alpha(R_{t+1} + \gamma Q(S_{t+1}, A_{t+1}) - Q(S_t, A_t))
            \end{split}
        \end{equation}
        where $\alpha$ is the learning rate parameter.
        
        So we first need to pick an action $a$ following the current policy using \(A_t = arg\,max_{a \in A}Q(S_t, a)\).
        By taking the action $A_t$, we observe a reward $R_{t+1}$ and get to the next state  $S_{t+1}$. We can then update our Q-value using :
        \[Q(S_t, A_t) \leftarrow Q(S_t, A_t) + \alpha(R_{t+1} +  \gamma \,max_{a \in A}Q(S_{t+1}, a) -  Q(S_t, A_t))\]
        
        \paragraph{$\epsilon$-greedy policy}
        In Reinforcement Learning, the exploration vs exploitation dilemma is crucial. Indeed, during the training, our agent needs not to get stuck in his current policy and continue exploring new actions but it also needs to use his policy more as its confidence grows.
        The $\epsilon$-greedy method offers a solution to this problem by picking a random action instead of using the current policy if a uniformly picked number is lower than an $\epsilon$ number. At the beginning of the training, $\epsilon$ is set to 1 so our agent does only exploration. As the training advances (and our agent gets more confident), we reduce $\epsilon$ so that our agent uses more his policy. Q-Learning is an off-policy method because it uses \(\,maxQ_{a \in A}(S_{t+1}, a)\) in his Q-value update no matter if the current $\epsilon$-greedy would have picked a random action or not. We use an $\epsilon$-greedy policy during the training of our models with $\epsilon$ starting from 1 and decreasing linearly to 0.1.
        
        \begin{figure}[hbt!]
            \centering
            \includegraphics[width=\linewidth]{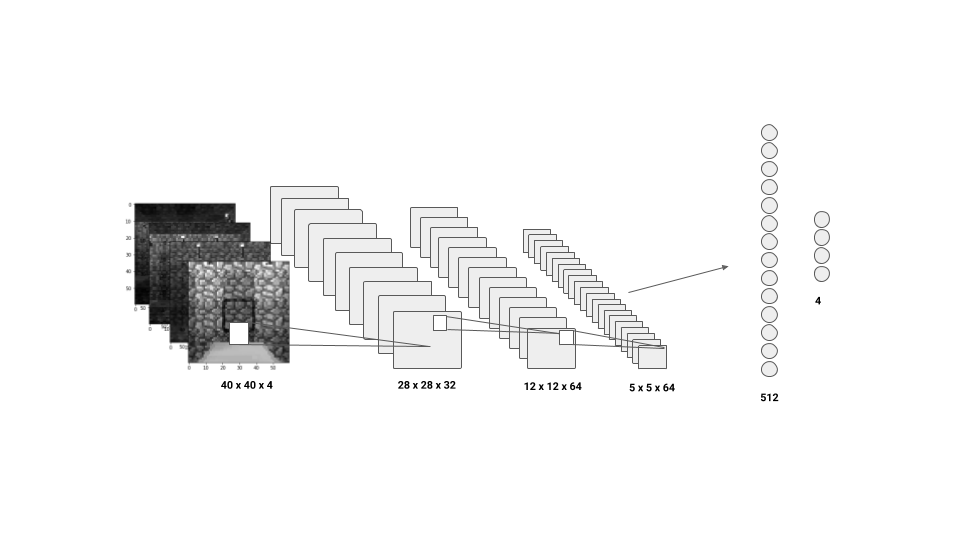}
            \caption{Our DQN architecture which takes the last four greyscaled frames and outputs the estimated Q-values. We used three convolutional layers each followed by a ReLu function and then a feedforward hidden layer with 512 neurons before the output layer.}
            \label{fig:dqn_architecture}
        \end{figure}
    
    \subsection{Deep Q-Learning}
        As previously mentioned, calculating the Q-value of every state-action pair gets computationally infeasible as the complexity of the environment grows (and so the state and action spaces). To tackle this problem, we can use a machine learning model to act as a function approximator of the Q-value. In the context of Deep Q-Learning, we use a neural network whose weights and biases are denoted $\theta$ in the new Q-value $Q(s, a; \theta)$. 
        
        Although several attempts had already been made, \citet{mnih_playing_atari_2013} introduced the Deep Q-Learning and reached human-level performances on several Atari 2600 games. Their architecture used a two layers convolutional neural network that takes as input the last four frames (in greyscale) stacked. They then used one feedforward hidden layer before the last layer outputting the Q-values. We use a similar architecture which can be seen in figure \ref{fig:dqn_architecture}. One can notice that these recent works are based on the first association of deep architectures and Q-learning \citep{lange_deep_2010} which followed the work of \citet{hutchison_neural_2005} who previously established a multi-layer perceptron dedicated to Q-learning.
        
        \paragraph{Experience replay}
        One of the key concepts brought by  \citet{mnih_playing_atari_2013} work is the use of experience replay. Each step is stored as an experience \(e_t = (s_t, a_t, r_t, s_{t+1})\) in a replay memory. Mini-batches of experiences are then randomly sampled and used to train the DQN. This allows the network to see multiple times each experience and removes the correlation between these samples. It was shown that this mechanism improves the stability of the training.
        
        \paragraph{Double Deep Q-Network}
        The DQN uses the loss function \(L(\theta) = \mathbb{E}_{(s, a, r, s') \sim U(D)}[(y - Q_{\theta}(s, a)^2]\) to update its weights $\theta$ using $s, a, r, s'$ sampled from the uniform distribution $U(D)$ over the experience replay memory. In order to calculate our target $y$, standard Q-Learning with function approximator would use \(y = r + \gamma \max_{a}Q(s_{t+1}, a)\). This method suffers from instability and to solve this, the extended version of the DQN \citep{mnih_human-level_2015} introduced a target network with weights $\theta^-$. This network is the same as the the original one except that its parameters are kept fixed and updated every $\tau$ steps to $\theta^-_t = \theta_t$. \citet{van_hasselt_deep_2015} improved this idea by adding the concept of Double Q-Learning \citep{hasselt_double_2010}. This technique is known to reduce the fact that the Q-Learning algorithm tends to overestimate Q-values. The calculus of $y$ now uses the original network to choose the next action but uses the target network to estimate the Q-value :
        \[y = r + \gamma Q_{\theta^-}(s_{t+1}, arg\,max_{a}Q_{\theta}(s_{t+1}, a))\]
        
        \begin{figure}[hbt!]
            \centering
            \includegraphics[scale=0.35]{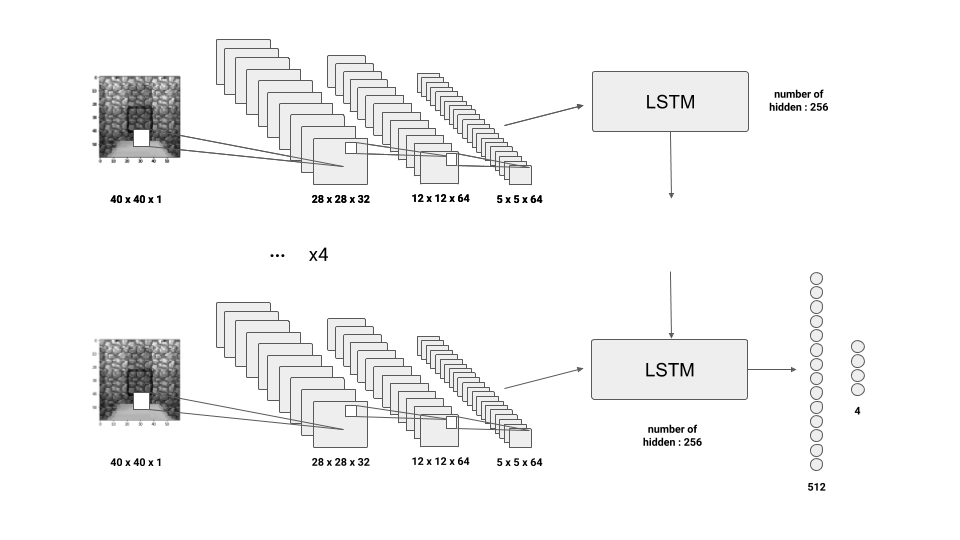}
            \caption{Our DRQN architecture which has an LSTM layer between the convolutional and the feedforward part of our DQN. The DRQN thus only takes the current frame as input and the LSTM layer uses the last hidden state to handle past dependencies. The hidden state is reset to zero at the beginning of every episode. Note that in order to accelerate the training, we give our DRQN a sequence of four frames during each gradient descent update.}
            \label{fig:drqn_architecture}
        \end{figure}
    
    \subsection{Deep Recurrent Q-Learning}
        The Deep Q-Network has shown great results but the main idea behind its architecture considers that the observation received contains all the information about the current state of the environment. In the context of 3D video games such as Minecraft, the agent only sees a fraction of the whole environment. \citet{mnih_human-level_2015} bypassed this issue by using the last four frames as input of their DQN and thus allowing the model to access more information than just the current observation. This solution works well for short dependencies needs\textemdash four in the case of \citet{mnih_human-level_2015}'s DQN\textemdash but can't work in environments where the agent needs to remember older information.
        
        To deal with such environments, \citet{hausknecht_deep_2015} modified the DQN architecture by adding a recurrent layer between the convolutional and the feedforward layers. Their model, called Deep Recurrent Q-Network (DRQN), now estimates \(Q(o_t, a_t, h_{t-1})\) instead of \(Q(s_t, a_t)\) where $o_t$ is the current observation (note that it's no longer $s_t$ which was considered as the whole current state) and $h_{t-1}$ is the hidden state of the agent at the previous step. They used LSTM \citep{hochreiter_long_1997} cells in the recurrent layer with $h_t =$ LSTM$(o_t, h_{t-1})$ and thus estimate \(Q(h_t, a_t)\). Our DRQN architecture is built on this method (see figure \ref{fig:drqn_architecture}).

        \begin{figure}[hbt!]
            \centering
            \includegraphics[width=\linewidth]{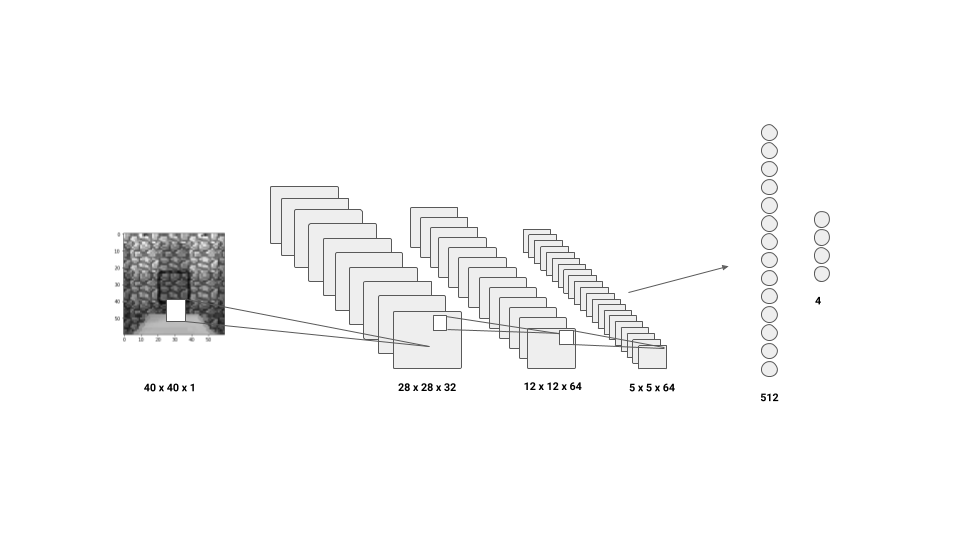}
            \caption{Our Simple DQN model. It has the same architecture as the DQN (see figure \ref{fig:dqn_architecture}) except that it only takes the current frame as input.}
            \label{fig:simple_dqn_architecture}
        \end{figure}
        
\section{Experiments}
    \subsection{Models}
        We chose to compare the performances of three Deep Q-Networks. They are all constituted of the same basis : A 6x6x32 convolutional layer, another 6x6x36 convolutional layer and a last 4x4x64 convolutional layer. Each convolutional layer is followed by a non-linearity function. The three networks also share the same architecture for the last two layers : A feedforward hidden layer with 512 neurons, and the output layer with four neurons with linear activation corresponding to the estimated Q-values. We used ReLu as both non-linearity function for the convolutional layers and activation function for the feedforward layer.
        
        We chose to downscale the frames observed from the environment to greyscale images in order to simplify and accelerate the training. We call our first tested network the Simple DQN (shown in figure \ref{fig:simple_dqn_architecture}). It takes the frame observed in input, pass it through the convolutional layers, flattens the feature maps and then gives it to the feedforward layers. This model has thus no structure intended to deal with the partial observability of our environment. Our second model, the DQN, is closer to the model introduced by \citet{mnih_playing_atari_2013}. It has the same structure as the previous one but takes the last four frames stacked as input in order to handle the short-term dependencies of our partially observable environment (see figure \ref{fig:dqn_architecture}). The last model, which we call the DRQN, has the same structure as the Simple DQN but has a recurrent layer before the feedforward hidden layer (see figure \ref{fig:drqn_architecture}). This recurrent layer is built with an LSTM cell with 256 units. The LSTM hidden state flows through the game episodes and is reset to zero at the beginning of every episode. We used four frames sequences for each mini-batch sample in order to train our DRQN.
        
        Note that all the models use the Double Deep Q-Learning method. We use an experience replay buffer from which we sample our mini-batches (of size 32 for the DQNs and 32/4 for the DRQN). We also use an $\epsilon$-greedy policy which starts from 1 and decreases linearly to 0.1 during the training. Moreover, we use a pre-training phase where our agent only plays randomly in order to fill the experience replay buffer. The details of our implementations can be found in appendix \ref{implementation_details}. For the training, we used a computer with a Intel Core I7 6700 processor, 16 Gb of RAM and a Nvidia GeForce GTX 1060 graphic card with 6 Gb. We used Tensorflow\footnote{https://www.tensorflow.org/} on the GPU of our graphic card.
        
    \subsection{Environments}
        We decided to compare our models performances on two simple missions. The first one is an easy version of the found-the-goal problem, where our agent is in a 7x7x7 room and has to find a block of gold (you can see a screenshot in figure \ref{fig:minecraft_overview}). Both the block and the agent spawn randomly on the $y$ axis and have a fixed position on the $x$ axis. In order to finish the mission, the agent must touch it. We restricted the maximum number of steps allowed to the agent to reach the goal to 40. A reward of 1 is given for the goal found and -1 if the maximum number of steps is reached. The agent also receives a -0.01 reward for each step taken.
        
        The other environment is the classical cliff walking problem where our agent starts from a point A and has to reach a point B without falling from the cliff. In our environment, our agent spawns on 8x3 pathway surrounded by lava. The objective is a block of diamond placed at the end of the pathway. The agent receives a reward of -0.01 for each step, 1 if it touches the goal and -1 if it drowned in lava or hasn't reached the goal after 70 steps. The length of 8 was chosen empirically. Indeed, it appeared during our tests that, with a length longer than 8, the random actions taken during the pre-training or with $\epsilon$-greedy policy couldn't get close enough to the objective. This caused our agents to never learn how to reach the goal and get stuck at 0\% of win. We believe our exploration strategy is not the best for this environment and are thinking about trying other methods (e.g.\ Thompson Sampling). Refer to section \ref{future_work} for more information. Also note that this is a very simple cliff walking problem and that there's no randomness in the generation of the cliff. As aforementioned, this work aims to check if the DRQN is always the best architecture on POMDPs when they are very simple. We thus made our cliff walking very simple in order to see if the DRQN could learn faster than the Simple DQN or the DQN.
        
        We restricted our agent's possible actions to :
        \begin{itemize}
            \item Move one step forward
            \item Move one step backward
            \item Turn head 90$\degree$ to right
            \item Turn head 90$\degree$ to left
        \end{itemize}
    
    \subsection{Results}
        To measure the performance of each model on the two environments, we monitored the training and then evaluated the trained models. The training performances rely mostly on three indicators measured every 50 episodes : the percentage of wins on these last 50 episodes, the mean of the number of steps that the agent took to make these wins and the mean of the accumulated rewards per episode. We also plotted the model's loss and the approximated Q-value (the network output) versus the target Q-value (the target from which the loss is calculated). These last two metrics can be found in appendix \ref{other_training_metrics}.
        
        Concerning the evaluation, we evaluated our trained models at three checkpoints of their training (in terms of episodes) because we thought it would give an interesting information about how fast the model learns. We made each evaluation three times on 100 episodes (with no seed fixed for the placement of the goal of the Basic mission) and kept the mean of the metrics to have more accurate results. We measured the percentage of wins and the mean of the number of steps by win. Moreover, every 50 episodes, we printed for each step the frame received by the agent, the Q-values estimated and the action chosen. This allowed us to better understand how our agents were behaving. These visualizations can be found in the appendix \ref{agents_decisions_visu}.
        
        \begin{figure}[hbt!]
            \centering
            \includegraphics[width=\linewidth]{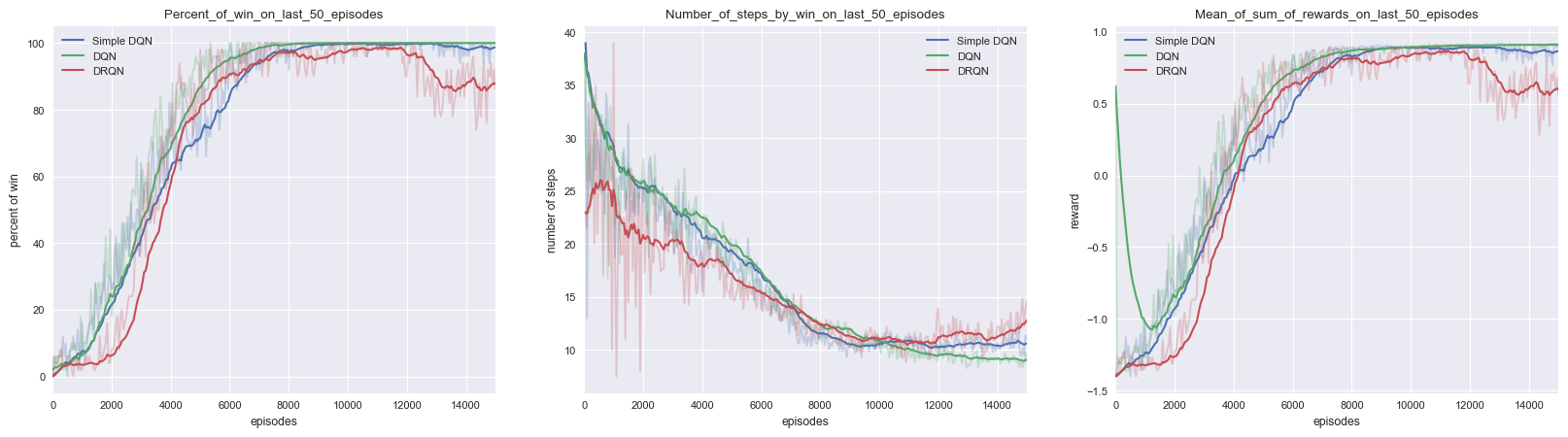}
            \caption{Training metrics on the Basic environment. We can see that the DQN is the fastest to converge and then stays at 100\% of victory whereas the Simple DQN slightly oscillates around 100\% at the end. We can also see that the DRQN clearly starts to get worse after 12\,000 episodes. }
            \label{fig:basic_graphs}
        \end{figure}
        
        \begin{table}[hbt!]
          \centering
          \setlength\tabcolsep{6pt}
          \begin{tabularx}{\textwidth}{@{}lcccccc@{}}
          \toprule
                   & \multicolumn{3}{c}{Percentage of wins (\%)} & \multicolumn{3}{c}{Nb steps by win (mean)} \\ \midrule
        Model      & 5000           & 10 000        & 15 000        & 5000              & 10 000            & 15 000            \\ \midrule
        Simple DQN & 39 ($\pm$ 3)   & 100 ($\pm$ 0) & 100 ($\pm$ 0) & 6.7 ($\pm$ 0.2)   & 9 ($\pm$ 0.1)     & 9.1 ($\pm$ 0.4)   \\
        DQN        & 100 ($\pm$ 0)  & 100 ($\pm$ 0) & 100 ($\pm$ 0) & 11.5 ($\pm$ 0.4)  & 9.1 ($\pm$ 0.2)   & 7.6 ($\pm$ 0.1)   \\
        DRQN       & 73 ($\pm$ 1)   & 97 ($\pm$ 2)  & 90 ($\pm$ 2)  & 8.9 ($\pm$ 0.2)   & 9.4 ($\pm$ 0.2)   & 12.5 ($\pm$ 0.5)  \\ \bottomrule
          \end{tabularx}
          \caption{Comparison of the models' evaluation results on the Basic environment. Though both the Simple DQN and the DQN reached 100\%  of win after 15\,000 episodes, the DQN clearly uses less steps and had already reached 100\% of win after only 5000 episodes. The DRQN seemed longer to train but floored after 10\,000 episodes and even started to be less efficient.}
          \label{tab:basic_results_table}
        \end{table}

        \subsubsection{Basic Mission}
            We chose to evaluate our models on the Basic mission to have a baseline of the performances on a very easy task where our agent can't die. We saved our models after 5000, 10\,000 and 15\,000 episodes (which means around 288\,000 steps for the Simple DQN, 276\,000 for the DQN and 303\,000 for the DRQN at the end of the training). In terms of training time, the Simple DQN was, as expected, the fastest (23 hours to reach 15\,000 episodes). The DQN was the longest because of our implementation of its input which needs to be padded with zeros when the episode has not yet seen four frames and also because of the complexity of its experience replay buffer. It took 1 day and 8 hours for the DQN to be trained, which is roughly 39\% longer than the Simple DQN. The DRQN was a bit longer than the Simple DQN (1 day and 1 hour) because of its experience replay buffer and because of the recurrent layer that adds complexity to the training.
            
            During the training, the DQN quickly achieved a mean of 100\% victory (after 6500 episodes) and then kept this result. Although almost as fast, the Simple DQN which reached the mean 100\% victory after 8500 episodes, kept oscillating around 100\% and started to get worse at 12\,500 episodes. On the other hand, the DRQN never reached that mean of 100\%. It had a similar behaviour as the two other models but floored around 95\% after 8000 episodes and its performance started to get sensibly deteriorated at 12\,000 episodes. The three models had a similar behaviour concerning the number of steps by win. They main difference is that the DQN kept getting better after 10\,000 episodes whereas the Simple DQN and the DRQN floored or became worse. All these results can be found in figure \ref{fig:basic_graphs}.
            
            We then evaluated each model after 5000, 10\,000 and 15\,000 episodes. The results obtained are gathered in table \ref{tab:basic_results_table}. The results showed that the DQN was already able to perform in 100\% of the episodes after only 5000 training episodes. The DQN obtained an average of 7.6 steps to win after 15\,000 episodes of training, which is clearly better than the two other models. The simple DQN reached the 100\% performance after 10\,000 episodes but, as seen in the training metrics, floored at 9 steps per win. Finally, the evaluation of the DRQN confirms what we had seen during its training. Indeed the 5000 episodes training version of the DRQN performs better than the Simple DQN, but it also floored after 10\,000 episodes and clearly worsened after 15\,000 episodes (only 90\% win and a mean of 12.5 steps). 
            
            \begin{figure}[hbt!]
                \centering
                \includegraphics[width=\linewidth]{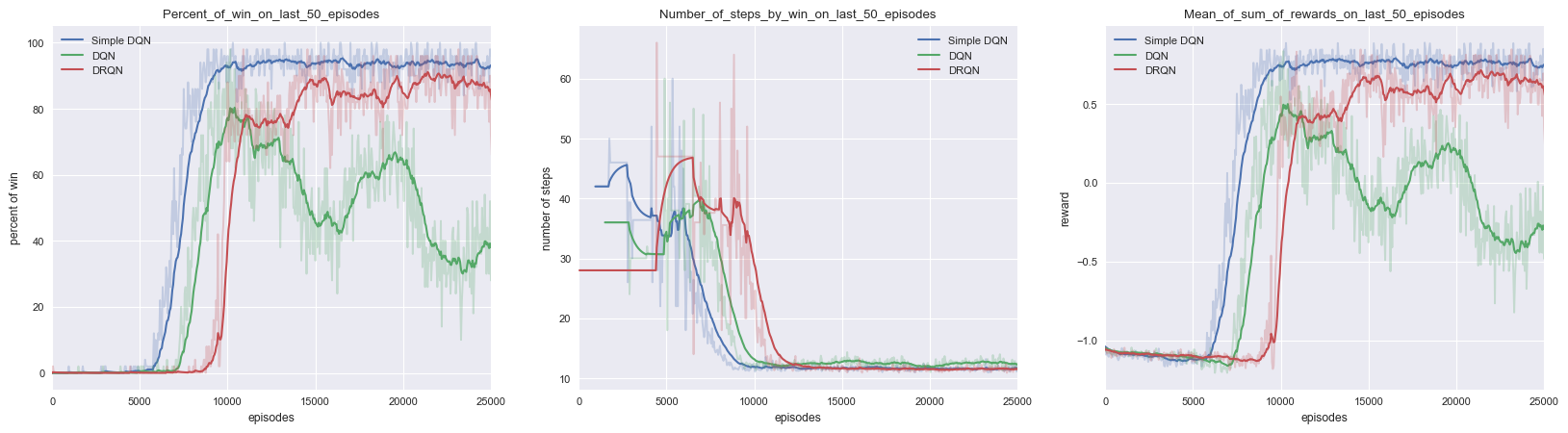}
                \caption{Training metrics on the Cliff Walking environment. These graphs show that the Simple DQN is clearly the fastest to converge and the most stable. The DQN had a similar behaviour to the Simple DQN at the beginning but stopped learning after 10\,000 episodes. The DRQN training was less smooth than the Simple DQN but it reached almost 90\% of win after 15\,000 episodes.}
                \label{fig:cliff_graphs}
            \end{figure}
            
            \begin{table}[hbt!]
              \centering
              \setlength\tabcolsep{6pt}
              \begin{tabularx}{\textwidth}{@{}lcccccc@{}}
              \toprule
                       & \multicolumn{3}{c}{Percentage of wins (\%)} & \multicolumn{3}{c}{Nb steps by win (mean)} \\ \midrule
            Model      & 10 000         & 20 000            & 25 000            & 10 000            & 20 000            & 25 000            \\ \midrule
            Simple DQN & 100 ($\pm$ 0)  & 99.7 ($\pm$ 0.3)  & 100 ($\pm$ 0)     & 10 ($\pm$ 0)      & 10 ($\pm$ 0)      & 10 ($\pm$ 0)      \\
            DQN        & 96 ($\pm$ 10)   & 83 ($\pm$ 9)     & 83 ($\pm$ 14)     & 10.1 ($\pm$ 0)    & 10.5 ($\pm$ 0.1)  & 10 ($\pm$ 0)      \\
            DRQN       & 93 ($\pm$ 3)   & 96.6 ($\pm$ 2)    & 99.3 ($\pm$ 0.7)  & 10 ($\pm$ 0.1)    & 9.4 ($\pm$ 0.2)   & 10 ($\pm$ 0)      \\ \bottomrule
              \end{tabularx}
              \caption{Comparison of the models' evaluation results on the Cliff Walking environment. As seen on the training graphs in figure \ref{fig:cliff_graphs}, the Simple DQN is capable of 100\% win with 10 steps after only 10\,000 episodes of training. On the other hand, the DQN floored and worsened. Finally, the DRQN slowly improved to almost reach 100\% after 25\,000 episodes. Note that the DRQN at 20\,000 episodes is the overall best model in terms of number of steps.}
              \label{tab:cliff_results_table}
            \end{table}
        \subsubsection{Cliff Walking Mission}
        After using the Basic environment, we wanted a little harder environment where our agent could die. We thus chose the Cliff Walking, with no randomness in the generation, where the task would still be easy to resolve even with the Simple DQN but maybe harder than the Basic. We trained our three networks on 25\,000 episodes and saved them after 10\,000, 20\,000 and 25\,000 episodes. After the end of training, the DRQN had seen 280\,000 frames, the Simple DQN 290\,000, and the DQN 236\,000. The whole training of the Simple DQN (25\,000 episodes) lasted 1 day and 4 hours. It was roughly the same for both the DQN (1 day and 3 hours) and the DRQN (1 day and 3 hours).
        
        As shown in figure \ref{fig:cliff_graphs}, the Simple DQN was the fastest to learn. It reached a mean of 90\% win after only 10\,000 episodes and then stayed around 95\%. The DRQN needed more episodes to reach his best training performance (around 15\,000 episodes to reach 90\%). It floored at around 90\% and then started to worsen after 25\,000 episodes. We were expecting the DQN to have the same behaviour as the Simple DQN but were surprised to see that it was the one struggling the most. It was converging like the two others when it stopped after 10\,000 episodes and then got worse.
        
        In the same way as the Basic environment, we evaluated our models trained on the Cliff Walking mission at three checkpoints : after 10\,000, 20\,000 and 25\,000 episodes. As expected with the shape of the training graphs in figure \ref{fig:cliff_graphs}, the Simple DQN was able to complete all the evaluations after only 10\,000 episodes of training. The DQN results were non consistent and the means of these results were getting worse with more trained versions. The best DRQN version (after 25\,000 episodes) succeeded 99.3\% of win but was still under the Simple DQN results.

\section{Conclusion}
    In this paper, we wanted to better understand the need of architecture modifications of the DQN in order to better deal with the partial observability of simple POMDPs. We thus tested models with no modification, stacked frames as input to handle short term past dependencies, and recurrent layers to handle long term dependencies. We used the 3D video game Minecraft, which can be framed as a POMDP, to design two simple and classical missions that could be solved by a vanilla DQN. 
    
    Our results on the Basic mission showed that the fact of stacking frames made the DQN much more efficient and faster than the Simple DQN on this mission. These results also showed that the DRQN could be tricky to train, longer, and not necessarily better. The DRQN showed better results on the Cliff Walking but still wasn't the best model. We were surprised that the Simple DQN was the fastest and most efficient model on that mission. The struggle of the DQN doesn't seem normal to us and requires further investigations to better understand this behaviour.
    
    These results on two very simple missions in a partially observable environment show that adding a recurrent layer to a DQN and thus making it harder and longer to train is not always the best option. The DRQN even showed worse performances than at least one of the other two models on both the Basic and the Cliff Walking missions. We believe that, when it comes to partially observable environments, one should wonder if the agent really needs to remember old information (it's for instance the case in hard exploratory problems such as the Obstacle Tower \citep{juliani_obstacle_2019}). If it is case, the vanilla DQN is clearly not suited to deal with this and needs to be modified. The DRQN could be one solution but, as aforementioned, others exist. If it is not the case, as it were in our two missions, the DQN can be enough depending on the complexity of the environment.

\section{Future Work} \label{future_work}
    This work acts as a baseline for us and opens perspectives. First, we believe our results can be improved. Secondly, we would like to test and improve the idea of DRQN on much harder partially observable environments. So, our future work will be focused on the four methods below.
    
    \paragraph{Dueling}
        One famous way to improve the DQN, is the \citet{wang_dueling_2015} Dueling DQN technique. This architecture is very close to our work since the only change is that we compute the Value function $ V(s)$ and the Advantage function $ A(a)$, both components of the Q-value, using two separate neural networks. By decoupling these estimations we could allow our agent to better understand them separately rather than understanding a total fusion of them. Indeed, calculating the Value and the Advantage together may be a problem if the current action doesn't influence a lot the state value or if we have many similar-valued actions. We could expect better empirical results implementing this technique.
    
    \paragraph{Thompson Sampling}
        \citet{azizzadenesheli_efficient_2018} demonstrated the naive aspect of the $\epsilon$-greedy exploration and proposed a Bayesian alternative using a Bayesian Deep-Q Network (BDQN). The idea is to use the Thompson Sampling technique through Gaussian sampling. Using Bayesian properties also allow uncertainty measurements for the Q-Values that may be interesting depending on the context. \citet{azizzadenesheli_efficient_2018} used Bayesian linear regression and Thompson Sampling which resulted in an efficient exploration giving higher rewards faster. As our agent is evolving in a POMDP, the exploration is very important, we could implement the Thompson Sampling hoping for a more efficient exploration.
    
    \paragraph{Policy Gradient}
        The recent methods generally obtaining the best results in Reinforcement Learning are the policy gradient algorithms. Unlike the off-policy methods, policy gradient techniques are policy-based which means that they try to directly find the best policy that maximizes the sum of rewards. The main current methods are the Trust Region Policy Optimization (TRPO) \citep{schulman_trust_2015} and the Proximal Policy Optimization (PPO)  \citep{schulman_proximal_2017}. The latter is the state-of-the-art algorithm as it's simpler and more efficient than its parent (TRPO). These algorithms are mainly used for 3D locomotion \citep{todorov_mujoco:_2012} because they suit continuous actions and high dimensions. Implementing one of these policy-based algorithms coupled with the two ideas aforementioned may improve significantly our results.
    
    \paragraph{Recurrent Neural Networks}
        As aforementioned in section \ref{related_work}, different approaches exist when it comes to add a recurrent part to the classical DQN. Indeed, \citet{oh_control_2016} showed that an approach with an external memory combined with a feedback connection could outperform both the DQN and the DRQN in a Minecraft environment. \citet{chen_deep_2016} and \citet{sorokin_deep_2015} also tried to add an Attention mechanism \citep{mnih_recurrent_2014} to help the recurrent layer to focus on interesting information of the past. They obtained promising results and we think there are still investigations to be done in this direction. Although the LSTM has been promising in his ability to handle long term dependencies, we believe different approaches could improve the performances of the DRQN we used in this paper.

\bibliographystyle{humannat}
\bibliography{bibliography}

\newpage

\begin{appendices}
    \section{Implementation Details} \label{implementation_details}
        In this section we present the details of our implementation. We list the hyperparameters we had to choose and the value we attributed to them. Note that they were chosen empirically throughout our tests. See table \ref{tab:general_imp_details} below.
        \input{appendices/implementation_details_table}
        
    \section{Other Training Metrics} \label{other_training_metrics}
        Apart from the percentage of wins, the mean of the number of steps by win and the mean of the accumulated rewards, we also used two other training metrics. The first one is the loss of the neural networks. The second is the difference between the Q-values estimated by the models and the target Q-values we give to our models, seen with a histogram from Tensorboard\footnote{\url{https://www.tensorflow.org/guide/summaries_and_tensorboard/}}. Remember that the loss of our models is \(L(\theta) = \mathbb{E}_{(s, a, r, s') \sim U(D)}[(y - Q_{\theta}(s, a)^2]\) with \(y = r + \gamma Q_{\theta^-}(s_{t+1}, \,max_{a}Q_{\theta}(s_{t+1}, a))\). Consequently the loss graph is the mean of the squared difference between the two Q-values histograms.
        
        \subsection{Basic Mission}
\begin{figure}[H]
                \centering
                \includegraphics[width=\linewidth]{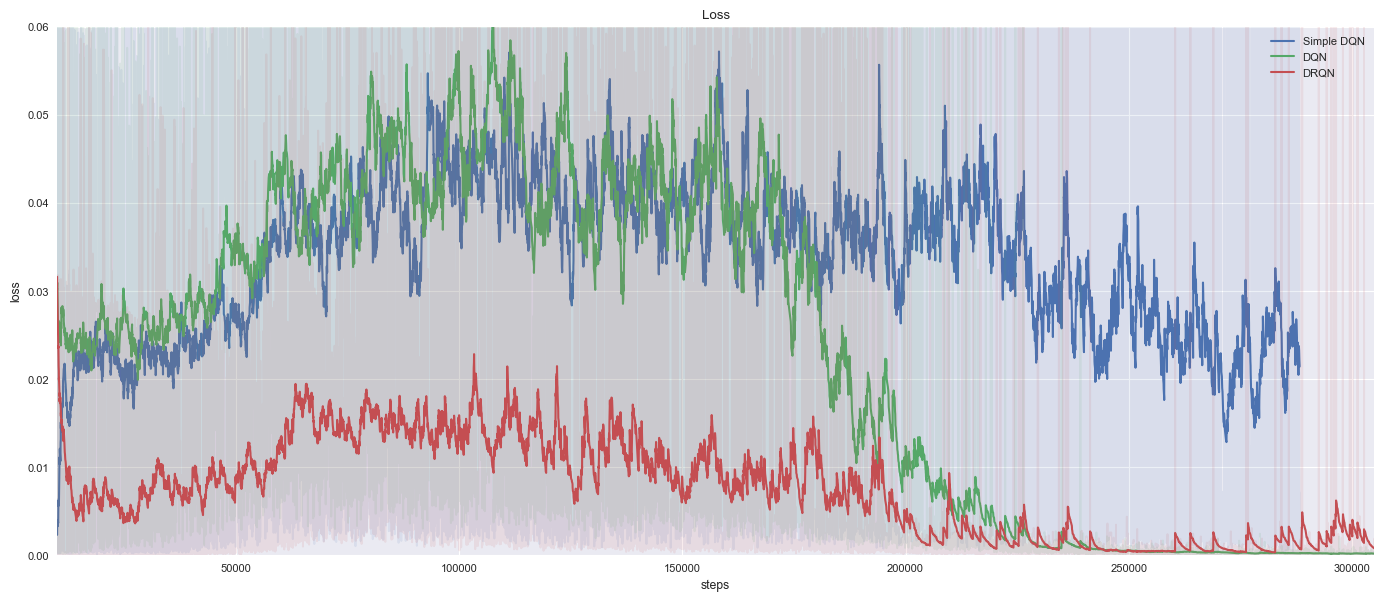}
                \caption{Losses of our three models on the Basic environment. The DQN reached 0 after 250\,000 steps, and though being the less efficient, the DRQN oscillates around 0 after 250\,000 steps too.}
                \label{fig:basic_loss_graphs}
            \end{figure}

\begin{figure}[H]
                \centering
                \includegraphics[width=\linewidth]{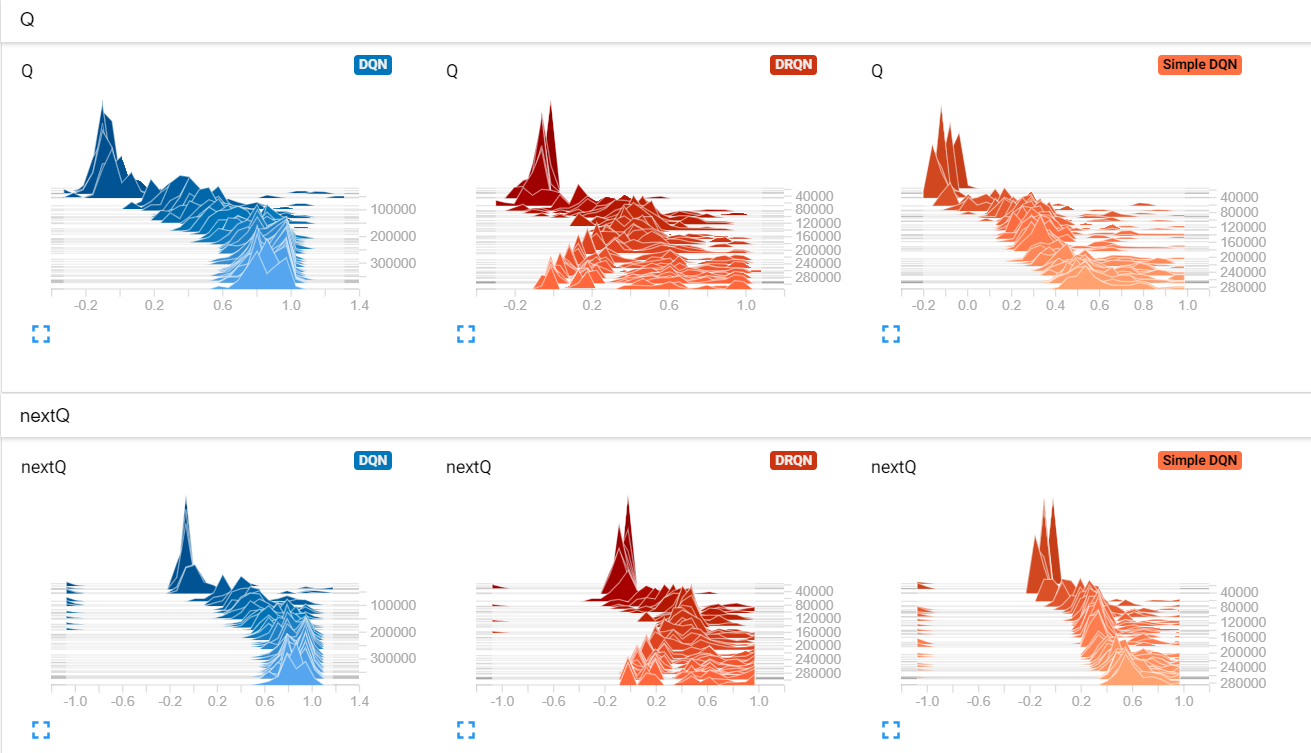}
                \caption{This is a screenshot of the histograms of the Q-values estimated by each model (top) and the target Q-value (bottom). Each graph plots the distribution of the Q-values of a mini-batch (remember that after the pre-training phase, we sample a mini-batch and train our model every 4 steps). We can see that the DQN is the most accurate (as seen in the losses in figure \ref{fig:basic_loss_graphs}).}
                \label{fig:basic_histo_graphs}
            \end{figure}

            You can see the losses of our three models on the Basic environment in figure \ref{fig:basic_loss_graphs} and the Q-values histograms on the Basic environment too in figure \ref{fig:basic_histo_graphs}.
            
        \subsection{Cliff Walking Mission}
\begin{figure}[H]
                \centering
                \includegraphics[width=\linewidth]{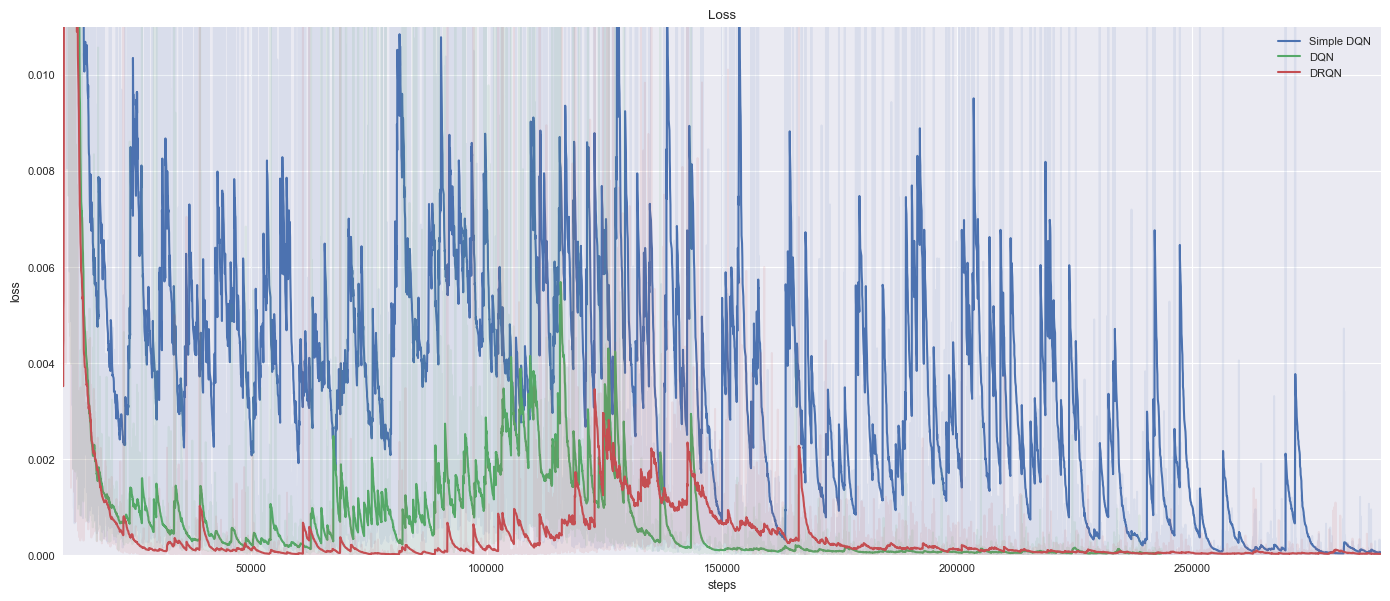}
                \caption{Losses of our three models on the Cliff Walking environment. As opposed to the results we obtained, we can see here that the Simple DQN was the longest to reduce his loss to 0.}
                \label{fig:cliff_loss_graphs}
            \end{figure}

\begin{figure}[H]
                \centering
                \includegraphics[width=\linewidth]{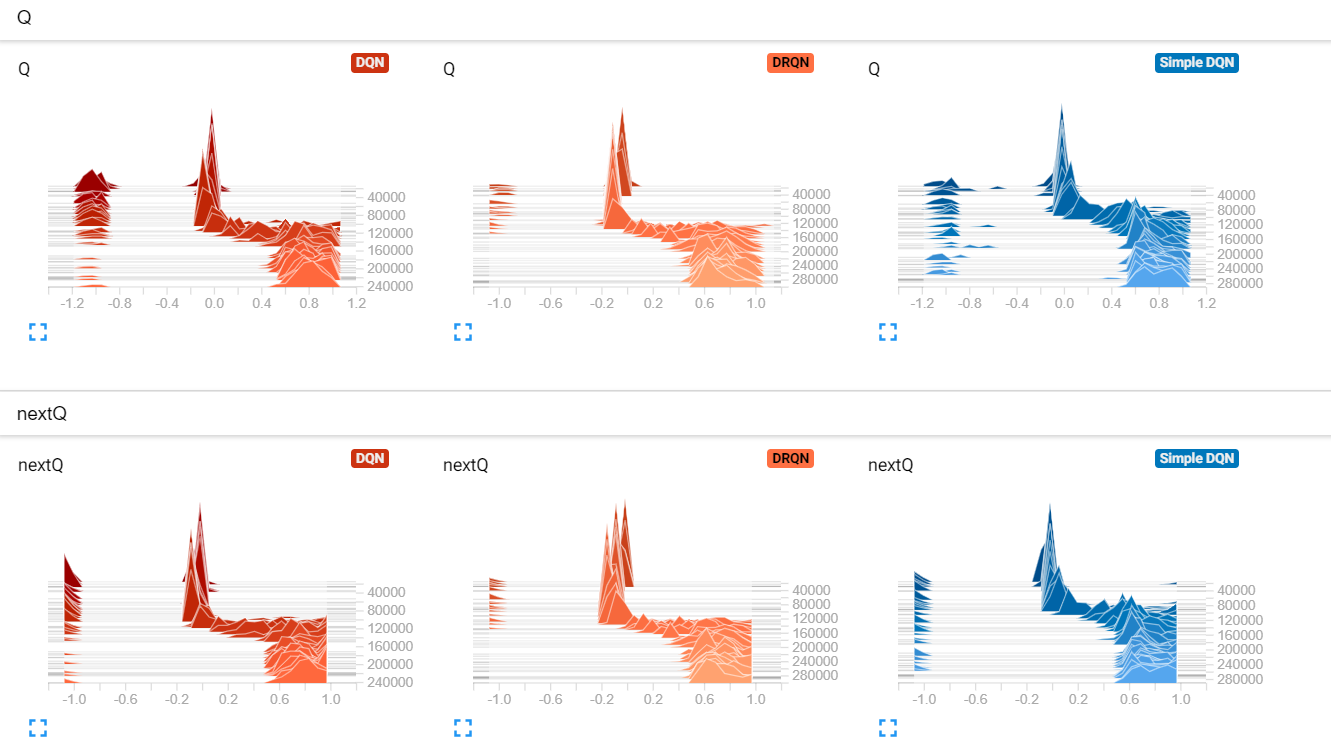}
                \caption{In this figure, we can see histograms of the Q-value estimated by the networks versus the target Q-values. All the networks seem to estimate well the Q-values.}
                \label{fig:cliff_histo_graphs}
            \end{figure}

        Here we show the training losses of our three networks and the evolution of the Q-values they estimate for the Cliff Walking environment in respectively figure \ref{fig:cliff_loss_graphs} and figure \ref{fig:cliff_histo_graphs}.
            
    \section{Agents Decisions Visualization} \label{agents_decisions_visu}
\begin{figure}[H]
            \centering
            \includegraphics[width=\linewidth]{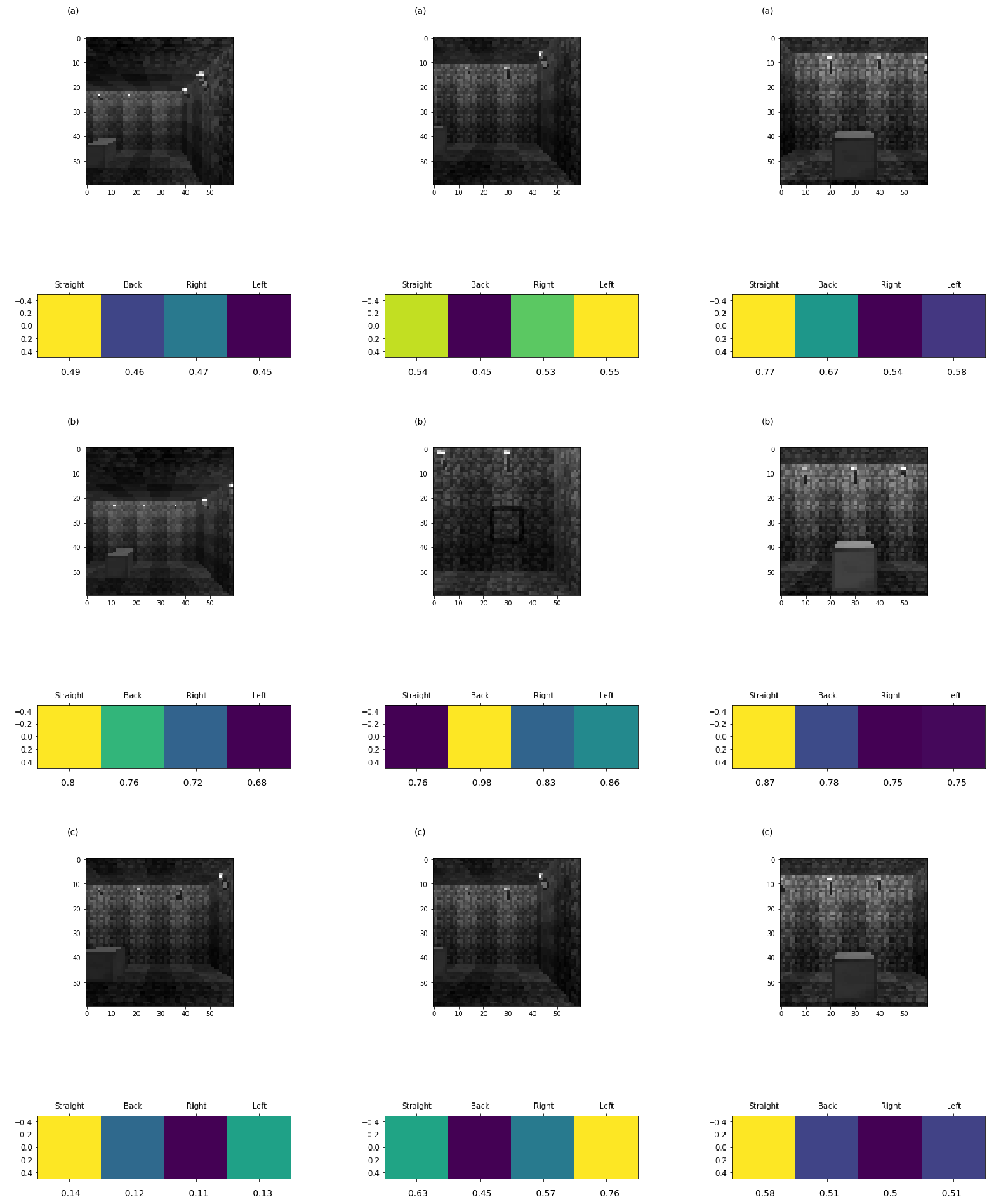}
            \caption{This figure shows the Q-values estimated by our models in three situations by model. We picked similar situations between models and remember that the final action taken was the max of the Q-values shown. The upper part corresponds to the Simple DQN model (a), the middle to the DQN (b), the bottom to the DRQN. We can see that the DRQN tends to underestimate the Q-values, and that the DQN seems more confident than the Simple DQN.}
            \label{fig:basic_decisions}
        \end{figure}

\begin{figure}[H]
            \centering
            \includegraphics[width=\linewidth]{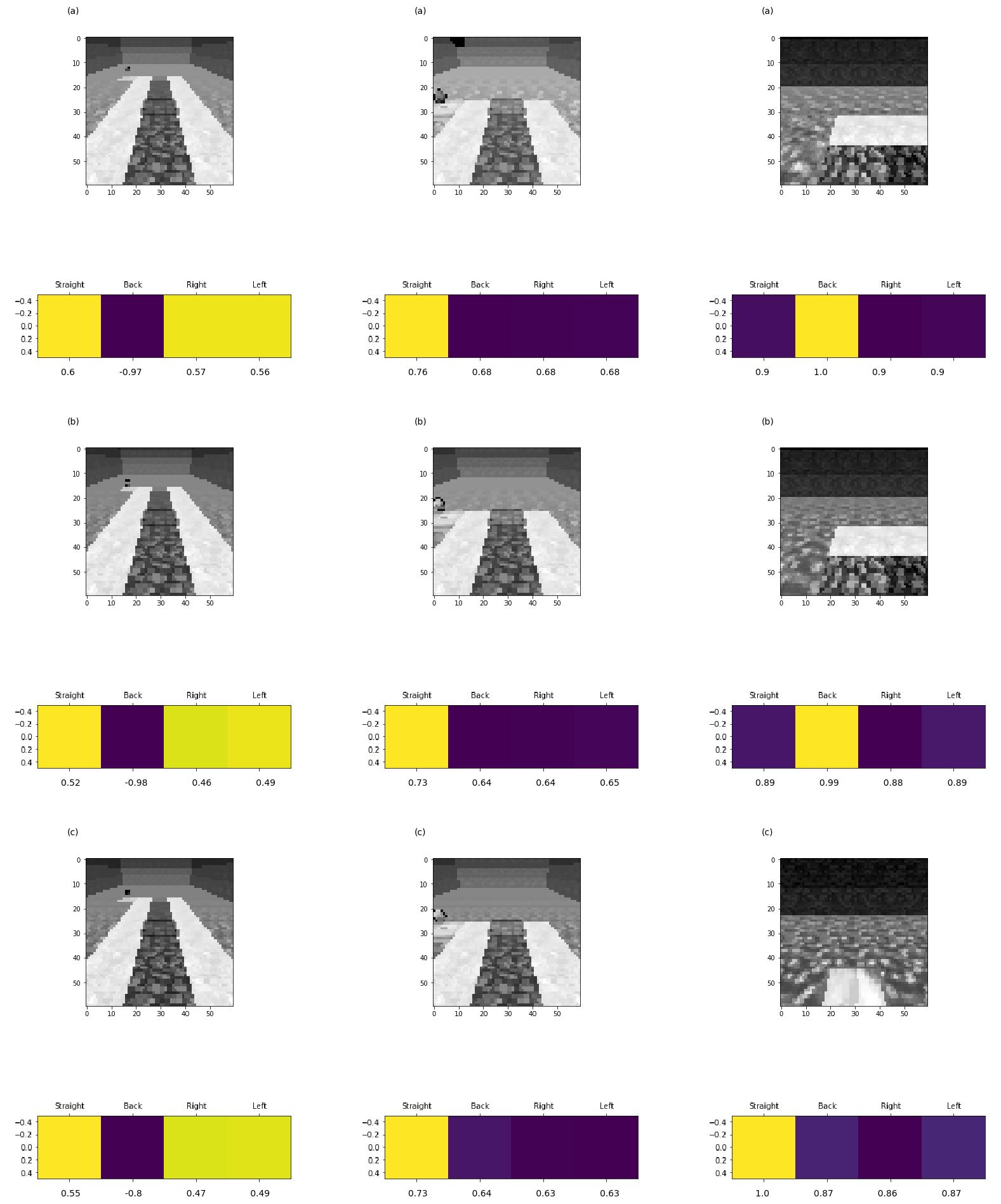}
            \caption{Here is an overview of the decisions taken by our agents in similar situations. The first line corresponds to the Simple DQN (a), the second one to the DQN (b) and the last one to the DRQN (c). There are no clear differences in the behaviour of our three agents on these situations except that the Simple DQN and DQN both learned to reach the goal by turning on the wrong side at the end (right instead of left) and reaching the goal by walking backward. This also appends on the Basic for the Simple DQN but we believe this is not related to the agent but to the exploration.}
            \label{fig:cliff_decisions}
        \end{figure}

        In order to better understand our agents' behaviour, we printed the frames received and the decisions taken for an episode every 50 episodes during the evaluation. We chose to extract three situations by agent (and tried to get similar situations between the agents) to illustrate this metric in this paper. We used only the final version of our models, so after 15\,000 episodes for the Basic and 25\,000 for the Cliff Walking. The decisions taken by our agents on the Basic environment can be found in figure \ref{fig:basic_decisions} whereas the ones for the Cliff Walking can be found in figure \ref{fig:cliff_decisions}.
        
\end{appendices}
\end{document}

%% file: appendices/implementation_details_table.tex
{\renewcommand{\arraystretch}{2}%
\begin{longtable}{llX}
    \caption{Hyperparameters used in our implementation.} \label{tab:general_imp_details} \\

    Hyperparameter  & Value & Description \\ \hline
    \endfirsthead

    \multicolumn{3}{c}{\tablename\ \thetable{} -- continued from previous page} \\
    Hyperparameter  & Value & Description \\ \hline
    \endhead
    
    \hline \multicolumn{3}{r}{{Continued on next page}} \\
    \endfoot
    
    \hline
    \endlastfoot

    input size & 60*60    & Size of the frames received by our agents. The frame is originally in RGB but we downscaled it to greyscale and then normalized it so each pixel value is between 0 and 1.         \\
    episode replay memory & 200 000 (steps) & Size of the experience replay buffer used for the Simple DQN. When this length is reached, we start popping elements from the beginning. \\
                          & 5 000 (episodes) & Size of the experience replay buffer used for the DQN. Note that this buffer stores episodes. \\
                          & 5 000 (episodes) & Size of the experience replay buffer used for the DRQN. Note that this buffer stores episodes.\\
    update frequency & 4 & Number of actions between two successive SGD updates. \\
    target network update parameter ($\tau$) & 0.001 & Rate parameter used to update the target Q-Network towards the original Q-network every mini-batch update. The equation is $\theta^- = \tau*\theta^- + (1-\tau)*\theta$.  \\
    mini-batch size & 32 & Size of the mini-batch given to the network for an SGD update for the Simple DQN and the DQN. They are sampled from the experience replay memory. \\
    & 8 & Size of the mini-batch given to the network for an SGD update for the DRQN. Corresponds to $original\_mini\_batch / agent\_history$. \\
    agent history length & 4 & History used by the agent. Corresponds to the number of frames stacked in input of the DQN or to the size of the sequence used by the DRQN during its training.  \\
    learning rate & 0.0001 & Learning rate of the Adam optimizer. \\
    number of LSTM units & 256 & Number of hidden units contained in the LSTM cells used by the DRQN. \\
    pre-training steps & 10 000 & Number of random steps done at the beginning of the training before starting to use the Deep Q-Network with the $\epsilon$-greedy policy. This is for the Basic environment. \\
    & 10 000 & Number of pre-training steps for the Cliff Walking environment. \\
    initial exploration & 1 & Initial value of $\epsilon$ \\
    final exploration & 0.1 & Initial value of $\epsilon$ \\
    annealing steps & 200 000 & Number of steps over which the initial value of $\epsilon$ is linearly annealed to its final value. This is the value used for the Basic environment.\\
    & 100 000 & Number of annealing steps for the Cliff Walking environment. \\
    discount factor & 0.99 & Discount factor $\gamma$ used in the Q-Learning algorithm. \\
\end{longtable}}